\documentclass[runningheads]{llncs}
\usepackage{amsmath}
\usepackage{amsfonts}
\usepackage{amssymb}
\usepackage{graphicx}
\usepackage[ruled]{algorithm2e}
\newcommand*\Laplace{\mathop{}\!\mathbin\bigtriangleup}

\usepackage{color}

\newcommand{\repeatthanks}{\textsuperscript{\thefootnote}}

\begin{document}
\title{Deep Patch-based Human Segmentation}
%
%
\author{Dongbo Zhang\inst{1}\thanks{Joint first author}
\and
Zheng Fang\inst{2}\repeatthanks
\and
Xuequan Lu\inst{3}
\and
Hong Qin\inst{4}
\and
Antonio Robles-Kelly\inst{3}
\and
Chao Zhang\inst{5}
\and
Ying He\inst{2}
}
\authorrunning{D. Zhang et al.}
%
\institute{Beihang University, China\\
\and
Nanyang Technological University, Singapore\\
\and
Deakin University, Australia\\
\and
Stony Brook University, USA\\
\and
University of Fukui, Japan
}
\maketitle              
\begin{abstract}
3D human segmentation has seen noticeable progress 
in recent years. It, however, still remains a challenge to date.
In this paper, we introduce a deep patch-based method for 3D human segmentation. We first extract a local surface patch for each vertex and then parameterize it into a 2D grid (or image). 
We then embed identified shape descriptors 
into the 2D grids which are further fed into the powerful 2D Convolutional Neural Network for regressing corresponding semantic labels (e.g., head, torso). Experiments demonstrate that our method is effective in human segmentation, and achieves state-of-the-art accuracy. 
\end{abstract}
%
%
%

\section{Introduction}
\label{sec:introduction}
3D human segmentation is a fundamental problem in human-centered computing. It can serve many other applications such as skeleton extraction, editing, interaction etc,. Given that traditional optimization methods have limited segmentation outcomes, deep learning techniques have been put forwarded to achieve better results.

Recently, a variety of 
human segmentation methods based upon deep learning have emerged \cite{guo20153d,haim2019surface,maron2017convolutional,masci2015geodesic}. The main 
challenges are twofold. Firstly the ``parameterization'' scheme and, secondly, the feature information as input. Regarding the parametrization scheme, some methods convert 3D geometry data to 2D image style with brute force \cite{guo20153d}. Methods such as \cite{maron2017convolutional} convert the whole human model into an image-style 2D domain using geometric parameterization. However, it usually requires certain prior knowledge like the selection of different groups of triplet points. Some methods like \cite{masci2015geodesic} simply perform a geodesic polar map. Nevertheless, such methods often need augmentation to mitigate origin ambiguity and sometimes generate poor patches for non-rigid humans. Regarding the input feature information, one simple solution is using 3D coordinates for learning which highly relies on data augmentation \cite{haim2019surface}. Other methods \cite{guo20153d,maron2017convolutional} employ shape descriptors like WKS \cite{aubry2011wave} as their input.

In this paper, we propose a novel deep learning approach for 3D human segmentation. In particular, we first cast the 3D-2D mapping as a geometric parameterization problem. We then convert each local patch into a 2D grid. We do this so as to embed both global features and local features into the channels of the 2D grids which are taken as input for powerful image-based deep convolutional neural networks like VGG \cite{simonyan2014very}. In the testing phase, we first parameterize a new 3D human shape in the same way as training, and then feed the generated 2D grids into the trained model to output the labels. 

We conduct experiments to validate our method and compare it with state-of-the-art human segmentation methods. Experimental results demonstrate that it achieves highly competitive accuracy for 3D human segmentation. We also conduct further ablation studies on different features and different neural networks.

\section{Related Work}

\subsection{Surface Mapping}
Surface mapping approaches solve the mapping or parameterization, ranging from local patch-like surfaces to global shapes. The Exponential Map is often used to parameterize a local region around a central point. It defines a bijection in the local region and preserves the distance with low distortion. Geodesic Polar Map (GPM) describes the Exponential Map using polar coordinates. \cite{floater2003mean,ju2005mean,schmidt2006interactive,melvaer2012geodesic} implemented GPM on triangular meshes based on approximate geodesics. Exact discrete geodesic algorithms such as \cite{surazhsky2005fast,xin2009improving} are featured with relatively accurate tracing of geodesic paths and hence polar angles. The common problem with GPM is that it easily fails to generate a one-to-one map due to the poor approximation of geodesic distances and the miscalculation of polar angles. To overcome the problem one needs to find the inward ray of geodesics mentioned in \cite{kokkinos2012intrinsic}. However, sometimes the local region does not form a topological disk and the tracing of the isocurve among the triangles is very difficult. To guarantee a one-to-one mapping in a local patch, one intuitive way is to adapt the harmonic maps or the angle-preserving conformal maps. A survey \cite{floater2005surface} reviewed the properties of these mappings. The harmonic maps minimize deformation and the algorithm is easy to implement on complex surfaces. However, as shown in \cite{duchamp1997hierarchical,floater1998parametric}, in the discrete context (i.e. a triangle mesh) if there are many obtuse triangles, the mapping could be flipped over. \cite{gu2003global,haker2000conformal,praun2003spherical,sheffer2001parameterization,sheffer2004robust} solved the harmonic maps on closed surfaces with zero genus, which is further extended to arbitrary-genus by \cite{gu2003global,khodakovsky2003globally}. These global shapes are mapped to simple surfaces with the same genus. If the domains are not homeomorphous, one needs to cut or merge pieces into another topology \cite{tutte1963draw,floater2003one}. These methods are globally injective and maintain the harmonicity while producing greater distortion around the cutting points.

\subsection{Deep Learning on Human Segmentation}
Inspired by current deep learning techniques, there have been a number of approaches attempting to extend these methods to handle the 3D human segmentation task. Limited by irregular domain of 3D surfaces, successful network architecture can not be applied straightforwardly. By leveraging Convolutional Neural Networks (CNNs), Guo et al. \cite{guo20153d} initially handled 3D mesh labeling/segmentation in a learning way. To use CNNs on 3D meshes, they reshape per-triangle hand-crafted features (e.g. Curvatures, PCA, spin image) into a regular gird where CNNs are well defined. 
This approach is simple and flexible for applying CNNs on 3D meshes. However, as the method only considers per-triangle information, it fails to aggregate information among nearby triangles which is crucial for human segmentation. At the same time, Masci et al. \cite{masci2015geodesic} designed the network architecture, named GCNN (Geodesic Convolutional Neural Networks), so as to deal with non-Euclidean manifolds. The convolution is based on a local system of geodesic polar coordinates to parameterize a local surface patch. This convolution requires to be insensitive to the origin of angular coordinates, which means it disregards patch orientation. Following \cite{masci2015geodesic}, anisotropic heat kernels were introduced in \cite{boscaini2016learning} to learn local descriptor with incorporating patch orientation. To use CNNs on surface setting, Maron et al. \cite{maron2017convolutional} introduced a deep learning method on 3D mesh models via parameterizating a surface to a canonical domain (2D domain) where the successful CNNs can be applied directly. However, their parameterization rely on the choice of three points on surfaces, which would involve significant angle and scale distortion. Later, an improved version of parameterization was employed in \cite{haim2019surface} to produce a low distortion coverage in the image domain. Recently, Rana et al. \cite{hanocka2019meshcnn} designed a specific method for triangle meshes by modifying traditional CNNs to operate on mesh edges. 

\section{Method}
\subsection{Overview}
\label{sec:overview}
In this work, we address 3D human segmentation by assigning a semantic label to each vertex with the aid of its local structure (patch).  Due to intrinsic irregularity of surfaces, traditional 2D CNNs can not be applied to this task immediately. To this end, we map a surface patch into a 2D grid (or image), in which we are able to leverage successful network architectures (e.g. ResNet \cite{he2016deep}, VGG \cite{simonyan2014very}). 

\begin{figure*}[htb]
    \centering
    \includegraphics[width=1.0\textwidth]{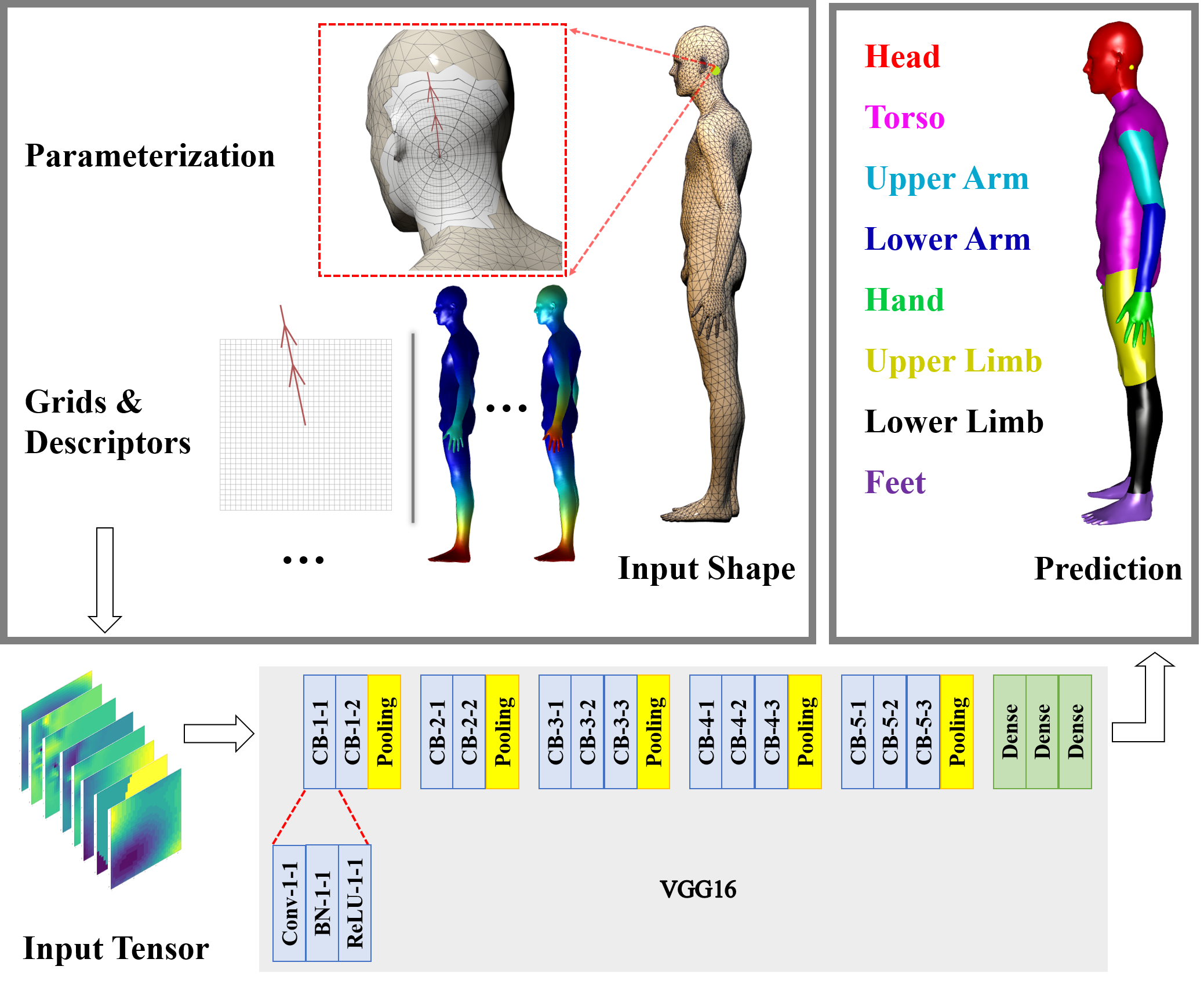}
    \caption{Overview of our method. 
    For each vertex, we first build a local patch on surface and then parameterize it into a 2D grid (or image). We embed the global and local features (WKS, Curvatures, AGD) into the 2D grid which is finally fed into VGG16 \cite{simonyan2014very} to regress its corresponding semantic label. }
    \label{fig:pipeline}
\end{figure*}

As shown in Fig. \ref{fig:pipeline}, for each vertex on a 3D human model, a local patch is built under geodesic measurement. We then convert each local patch into a 2D grid (or image) via a 3D-2D mapping step, to suit the powerful 2D CNNs. To preserve geometric information both locally and globally, we embed local and global shape descriptors into the 2D grid as input features. Finally, we establish the relation between per-vertex (or per-patch) feature tensor and its corresponding semantic label in a supervised learning manner. We first introduce the surface mapping step for converting a local patch into 2D grid in Section \ref{sec:dataprocessing}, and then explain the neural network and implementation details in Section \ref{sec:neuralnetworkandtrianingdetails}. 

\subsection{Surface Mapping}
\label{sec:dataprocessing}
\textbf{Patch extraction.} Given a triangular mesh $M$, we compute the local patch $P$ for each vertex $v\in M$ based on the discrete geodesic distance $d$ by satisfying $d(v_p)<r_p$ for all $v_p \in P$. $r_p$ is an empirically fixed radius for all patches. Assume the area of $M$ is $\alpha$, $r_p = \sqrt{(\alpha/m)}$, where $m$ is set to $1000$ in this work. The geodesic distance $d$ is computed locally using the ICH algorithm \cite{xin2009improving} due to its efficiency and effectiveness. 

\textbf{Parameterization.} There are two cases for parameterization in our context, whereby 
$P$ is a topological disk and otherwise. For the former case, we denote the 2D planar unit disk by $D$, we compute the harmonic maps $\mu: P\to D$ by solving the Laplace equations
\begin{equation}\label{eq:harmonicequ}
    \sum_{(v_j,v_i)\in M} c_{ij}(\mu(v_j)-\mu(v_i)) = 0,
\end{equation}
with Dirichlet boundary condition
\begin{equation}\label{eq:boundarycon}
\mu(v'_k) = (\cos\theta_k, \sin\theta_k), \theta_k = 2\pi \frac{\sum_{l=1}^k |v'_{l}-v'_{l-1}|}{\sum_{o=1}^m |v'_{o}-v'_{o-1}|},
\end{equation}
where $v_i$ is an interior vertex of $P$ (Eq. \eqref{eq:harmonicequ}) and $c_{ij}$ is the cotangent weight on edge $(v_i,v_j)$. In Eq. \eqref{eq:boundarycon}, $v'_k$ ($k\in[1,m]$) belongs to the boundary vertex set of $P$. The boundary vertex set contains $m$ vertices, which are sorted in a clockwise order according to the position on the boundary of $P$. Suppose $(v_i,v_j)$ is an interior edge. $(v_i,v_j,v_k)$ and $(v_i,v_j,v_l)$ are two adjacent triangles, $c_{ij}$ is calculated as
\begin{equation}
c_{ij} = \frac{1}{2}(\cot\beta_k + \cot\beta_l),
\end{equation}
where $\beta_k$ and $\beta_l$ is the angle between $(v_i,v_k)$ and $(v_j,v_k)$, and between $(v_i,v_l)$ and $(v_j,v_l)$, respectively.

\begin{figure*}[htb]
    \centering
    \includegraphics[width=1.0\textwidth]{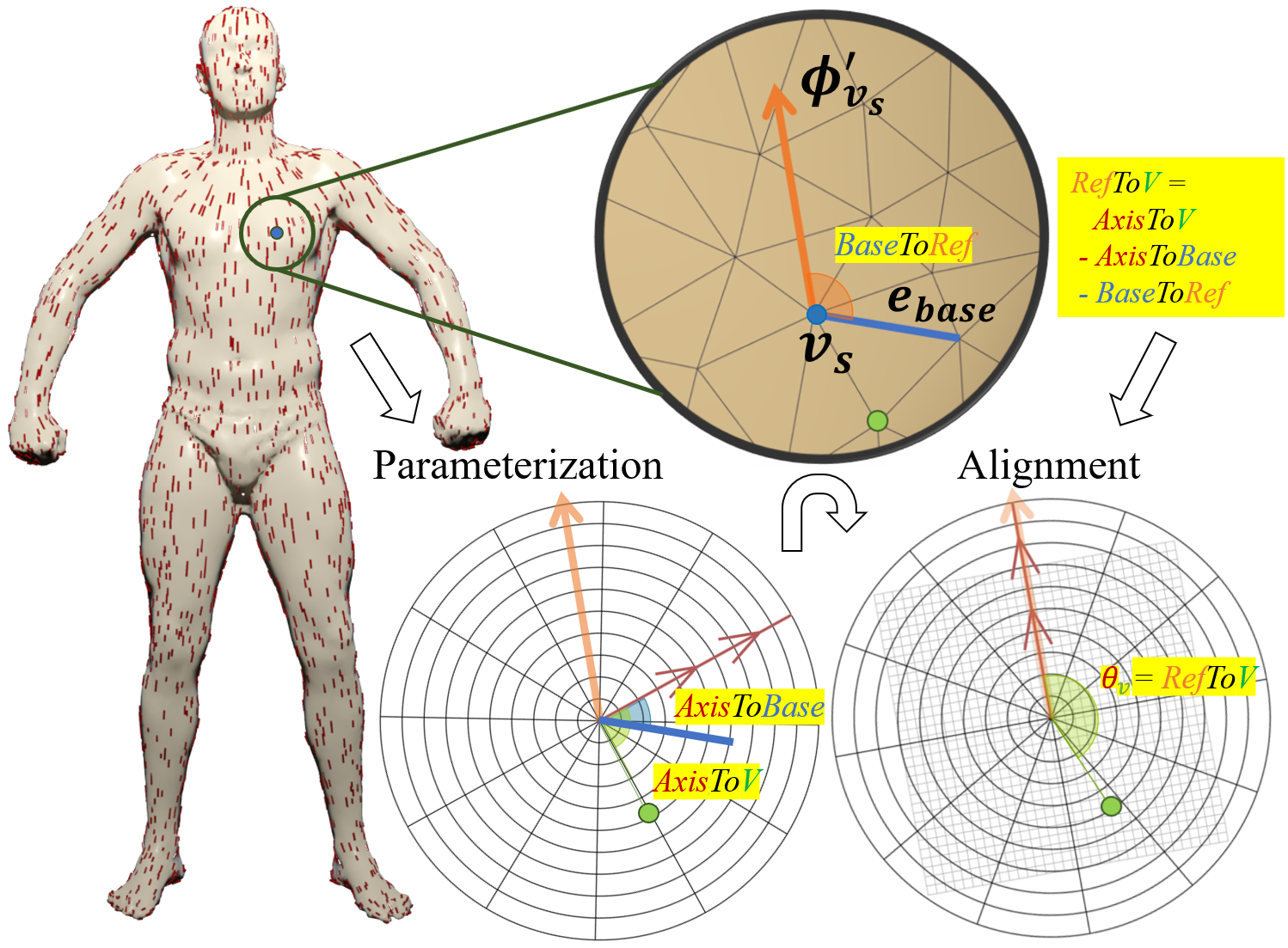}
    \caption{Left: flow vector field $\mathcal{\phi}$ rendered in red color on a human model. Top right: the close-up view of a local patch $P$ around $v_s$. We show the projected flow vector $\phi'(v_s)$ emanating from $v_s$ in orange and the base edge $e_{base}$ in blue. Bottom left: surface mapping. The red double-arrow line indicates the polar axis of the local polar coordinate system. Green dots represent the parameterized vertices from $P$. Bottom right: alignment of polar angles and grid discretization. The polar axis is rotated to overlap $\phi'(v_s)$. The angle of rotation is indicated in the yellow box. $\theta_v$ is the new polar angle for the vertex. After the alignment, the grid with $32\times32$ cells are embedded to the unit disk $D\in\mathbb{R}^2$.}
    \label{fig:parameterization}
\end{figure*}

There are cases where the local patch $P$ is not a topological disk and the harmonic maps can not be computed. In this case, we trace the geodesic paths for each $v_p\in P$, by reusing the routing information stored by the ICH algorithm when computing $d$. See Fig.~\ref{fig:parameterization} for illustration of parameterization. Similar to \cite{masci2015geodesic}, we then obtain a surface charting represented by polar coordinates on $D$. We next perform an alignment and a $32\times32$ grid discretization on $D$.

\textbf{Alignment and grid discretization.} The orientation of $D$ is ambiguous in the context of the local vertex indexing. We remove the ambiguity by aligning each patch with a flow vector field $\mathcal{\phi}$ on $M$. For each vertex $v \in M$ and its associated patch $P_v$, the flow vector $\mathcal{\phi}(v)$ serves as the reference direction of $P_v$ when mapping to $\mathbb{R}^2$. Fig.~\ref{fig:parameterization} illustrates the reference direction as an example. $\mathcal{\phi}$ is defined as 
a vector field flowing from a set of pre-determined sources $v_s \in M$ to the sinks $v_t \in M$. We initially solve a scalar function $u$ on $M$ using the following Laplace equation.
\begin{gather*}
    \Laplace u(v) = 0,\\
    u(v_s) = 0,~u(v_t) = 1,
\end{gather*}
and the flow vector field is $\mathcal{\phi} = \nabla u$.
We further calibrate the polar angles of $D$ with $\phi$. Considering the first adjacent edge $e_{base}$ around a source vertex $v_s$ as a base edge, \textit{BaseToRef} is the angle between the projected flow vector $\phi'(v_s)$ and $e_{base}$. $\phi'$ is the projected $\phi$ onto a random adjacent face of the base edge. From the harmonic maps $\mu$, we easily obtain the polar angles of the local, randomly-oriented polar coordinate system. The polar angles are represented by \textit{AxisToV} for all $v \in D$ and \textit{AxisToBase} for $e_{base}$. To align the local polar axis to the reference direction, the calibrated polar angle for all $v \in D$ is calculated as $\theta_v = \textit{AxisToV}-\textit{AxisToBase}-\textit{BaseToRef}$.

The grid with $32\times 32$ cells is embedded inside the calibrated $D$ such that $D$ is the circumcircle of the grid. We build a Cartesian coordinate system in $D$, and the origin is the pole in the polar system. The x-axis and the y-axis overlap the polar axis and $\theta = \pi/2$, respectively. The vertices and triangles on $D$ are converted to this Cartesian coordinate system. Some cells belong to a triangle if the cell centers are on the triangle. We compute the barycentric coordinates of each involved cell (center) with respect to the three vertices of that triangle. The barycentric coordinates will be used for calculating cell features based on vertex features later. 

\textbf{Shape descriptors.} 
After generating $32\times 32$ grids (or images), we embed shape descriptors as features into them. The features of each cell are calculated with linear interpolation using the barycentric coordinates computed above. The descriptors include Wave Kernel Signature (WKS)~\cite{aubry2011wave}, curvatures (minimal, maximal, mean, Gaussian) and average geodesic distance (AGD)~\cite{hilaga2001topology}. We normalize each kind of descriptors on a global basis, that is, the maximum and minimum values are selected from the descriptor matrix, rather than simply from a single row or column of the matrix.

\subsection{Neural Network and Implementation}
\label{sec:neuralnetworkandtrianingdetails}
\textbf{Neural network.} As a powerful and successful network, we adopt VGG network architecture with $16$ layers (see Fig. \ref{fig:pipeline}) as our backbone in this work. The cross-entropy loss is employed as our loss function for the VGG16 net. It is worth noting, however, that the surface parameterization presented in this work is quite general in nature, being applicable to many other CNNs elsewhere in the literature.

\textbf{Implementation details.}
We implement the VGG16 network in PyTorch on a desktop PC with an Intel Core i7-9800X CPU (3.80 GHz, 24GB memory). We set a training epoch number of $200$ and a mini-batch size of $64$. SGD is set as our optimizer and the learning rate is decreased from $1.0\times10^{-3}$ to $1.0\times10^{-9}$ with increasing epochs. To balance the distribution of each label, in the training stage we randomly sample $5,000$ samples per label in each epoch. Training takes about $3.5$ hours on a GeForce GTX 2080Ti GPU (11GB memory, CUDA 9.0).

Once the model is trained, we can infer semantic labels of a human shape in a vertex-wise way. Given a human shape, we first compute the involved shape descriptors for each vertex. 
For each vertex, we build a local surface patch and parameterize it into a 2D grid (or image) as described in Section \ref{sec:dataprocessing}. We embed all the shape descriptors into a 2D grid and feed it into our trained model for prediction. 

\section{Experimental Results}
In this section, we first introduce the dataset used in our experiments, 
and then explain the evaluation metric. We then show the visual and the quantitative results. We also perform ablation studies for the input features and different neural networks.

\subsection{Dataset Configuration}
\begin{figure*}[htb]
    \centering
    \includegraphics[width=1.0\textwidth]{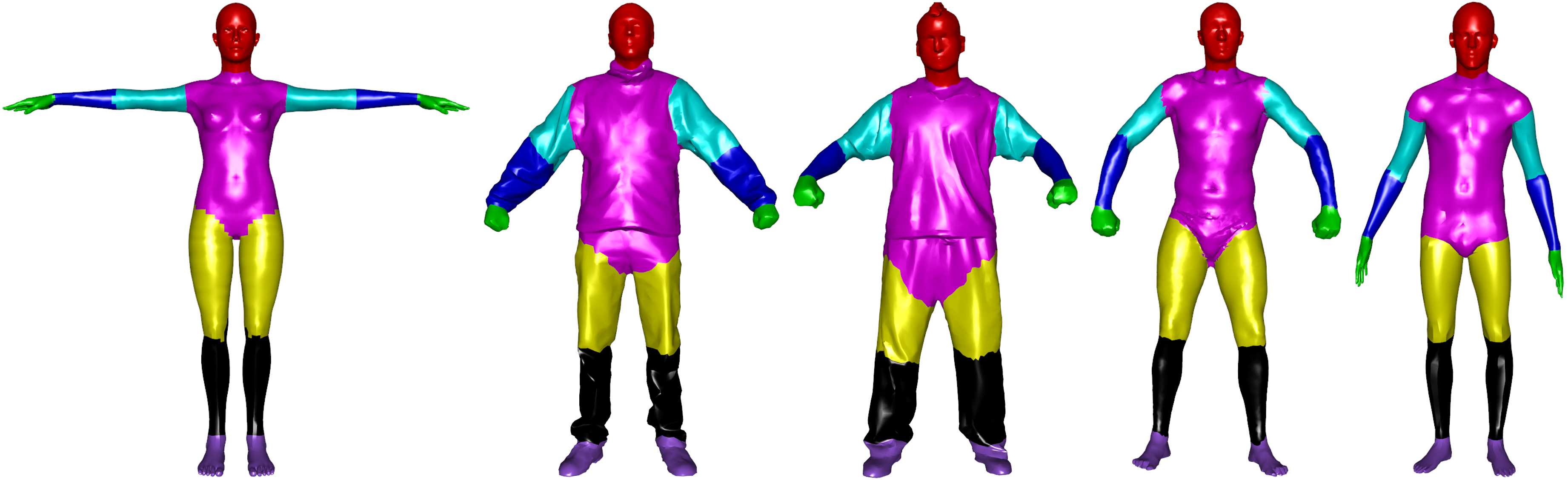}
    \caption{Examples from the the training set. }
    \label{fig:train_examples}
\end{figure*}

In this work, we use dataset from \cite{maron2017convolutional} which consists of $373$ train human models from SCAPE \cite{anguelov2005scape}, FAUST \cite{bogo2014faust}, MIT \cite{vlasic2008articulated} and Adobe Fuse \cite{AdobeFuse}, and $18$ test human models from SHREC07 \cite{giorgi2007shape}. Some examples of our training dataset are shown in Fig. \ref{fig:train_examples}. For each human model, there are $8$ semantic labels (e.g., Head, Arm, Torso, Limb, Feet), as shown in Fig. \ref{fig:pipeline}. To represent geometric information of a human model both  globally and locally, we concatenate a set of shape descriptors as input features: $26$ WKS features \cite{aubry2011wave}, $4$ curvature features ($C_{min}$, $C_{max}$, $C_{mean}$, $C_{gauss}$) and AGD \cite{hilaga2001topology}. 

\subsection{Evaluation Metric}
To provide a fair comparison, we also evaluate our segmentation results in an area-aware manner \cite{maron2017convolutional}. For each segmentation result, the accuracy is computed as a weighted ratio of correctly labeled triangles over the sum of all triangle area. Therefore, the overall accuracy on all involved human shapes is defined as
\begin{equation}\label{eq:acc}
ACC = \frac{1}{N} \sum_{i=1}^{N} \frac{1}{A_{i}} \sum_{j \in J_{i}} a_{ij},
\end{equation}
where $N$ denotes the number of test human models and $A_{i}$ is the sum of triangle area of the $i$-th human model.
$J_{i}$ is the set including the indices of correctly labeled triangles of the $i$-th human model and $a_{ij}$ represents the $j$-th triangle area of the $i$-th human model. Since we address the human segmentation task in a vertex-wise manner, the per-vertex labels need to be transferred into per-face labels for the quantitative evaluation. The face label is simply estimated by using a voting strategy among its three vertex labels. We immediately set the label with two or three vertices as the label on the face. We randomly select a vertex label as the face label, if three vertex labels are totally different.

\begin{figure*}[htb]
    \centering
    \includegraphics[width=1.0\textwidth]{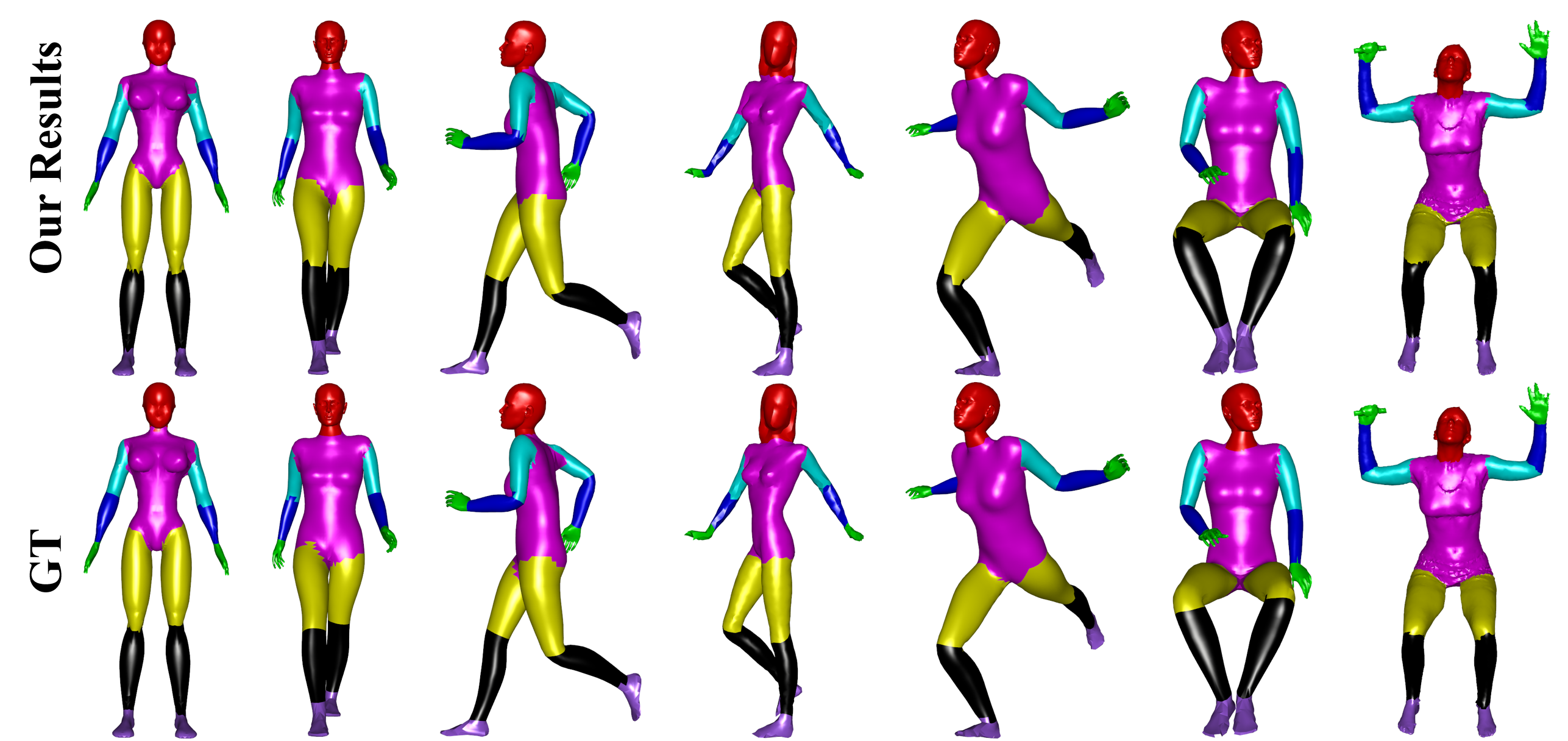}
    \caption{Some visual results of our method on the test set. The top row and the bottom row respectively show the results of our method and the corresponding ground truth models. }
    \label{fig:test_results}
\end{figure*}

\subsection{Visual and Quantitative Results}
In this section, we show the visual and quantitative results. As shown in Fig. \ref{fig:test_results}, the top row lists several of our results in the test set, and the bottom row displays the corresponding ground-truth models. To further evaluate our method for 3D human segmentation, a quantitative comparison with recent human segmentation techniques are summarized in Table \ref{tab:comparedresults}. As we can see from Table \ref{tab:comparedresults}, our method achieves an accuracy of $89.89\%$, ranking the second place among all methods. Our approach is a bit inferior to the best method \cite{haim2019surface} which certainly benefits from its data augmentation strategy.

\begin{table*}[htbp]
  \centering
  \caption{ Comparisons with recent methods for 3D human segmentation. } \label{tab:comparedresults}
  \setlength{\tabcolsep}{7mm}{
  \begin{tabular}{ccc}
  \hline
  Method & \#Features & ACC \\
  \hline
  DynGCNN \cite{wang2019dynamic} & 64 & 86.40\% \\
  Toric CNN \cite{maron2017convolutional} & 26 & 88.00\% \\
  MDGCNN \cite{poulenard2018multi} & 64 & 89.47\% \\
  SNGC \cite{haim2019surface} & 3 & 91.03\% \\
  GCNN \cite{masci2015geodesic} & 64 & 86.40\% \\
Our Method & 31 & 89.89\% \\
  \hline
  \end{tabular}}
\end{table*}%

\subsection{Ablation Study}
Besides the above results, we also evaluate different selection choices for input features. Table \ref{tab:ablationstudy_feats} shows that the input features including WKS, curvatures and AGD obtain the best performance, in terms of accuracy. Moreover, we evaluate the performance of two different neural networks in 3D human segmentation, as shown in Table \ref{tab:ablationstudy_network}. It is obvious that the VGG16 obtains a better accuracy than the ReseNet50, and we thus employ VGG16 as the backbone in this work. 

\begin{table*}[htbp]
  \centering
  \caption{Comparisons for different input features. For simplicity, S, W, C and A are respectively short for SI-HKS, WKS, Curvatures (Cmin, Cmax, Cmean, Cgauss) and AGD. } \label{tab:ablationstudy_feats}
  \setlength{\tabcolsep}{7mm}{
  \begin{tabular}{lcc}
  \hline
   Features Used & \#Features & ACC \\
  \hline
   SWCA & 50 & 89.25\% \\
   SWA  & 46 & 89.81\% \\
   WCA (Our) & 31 & 89.89\% \\
  \hline
  \end{tabular}}
\end{table*}%

\begin{table*}[htbp]
  \centering
  \caption{Comparisons for two different network architectures. } \label{tab:ablationstudy_network}
  \setlength{\tabcolsep}{7mm}{
  \begin{tabular}{lcc}
  \hline
   Network &  Features & ACC \\
  \hline
   ResNet50 & WKS, Curvatures, AGD & 87.60\% \\
   VGG16  & WKS, Curvatures, AGD & 89.89\% \\
  \hline
  \end{tabular}}
\end{table*}%


\section{Conclusion}
We have presented a deep learning method for 3D human segmentation. Given a 3D human mesh as input, we first parameterize each local patch in the shape into 2D image style, and feed it into the trained model for automatically predicting the label of each patch (i.e., vertex). Experiments demonstrate the effectiveness of our approach, and show that it can achieve state-of-the-art accuracy in 3D human segmentation. In the future, we would like to explore and design more powerful features for learning the complex relationship between the non-rigid 3D shapes and the semantic labels.

\bibliographystyle{splncs04}
\bibliography{mybibliography}

\end{document}